\pdfoutput=1

\documentclass[11pt]{article}

\usepackage[final]{acl}

\usepackage{times}
\usepackage{latexsym}

\usepackage[T1]{fontenc}

\usepackage[utf8]{inputenc}

\usepackage{microtype}

\usepackage{inconsolata}

\usepackage{graphicx}

\title{Significance of Chain of Thought in Gender Bias Mitigation for English-Dravidian Machine Translation}

\author{Lavanya Prahallad \\
  Research Spark Hub Inc., USA. \\
  \texttt{prahalladlavanya@gmail.com} \\\And
  Radhika Mamidi \\
  IIIT Hyderabad, India. \\
  \texttt{radhika@iiit.ac.in} \\}

\begin{document}
\maketitle
\begin{abstract}
Gender bias in machine translation (MT) systems poses a significant challenge to achieving accurate and inclusive translations. This paper examines gender bias in machine translation systems for languages such as Telugu and Kannada from the Dravidian family, analyzing how gender inflections affect translation accuracy and neutrality using Google Translate and ChatGPT. It finds that while plural forms can reduce bias, individual-centric sentences often maintain the bias due to historical stereotypes. The study evaluates the Chain of Thought processing, noting significant bias mitigation from 80\% to 4\% in Telugu and from 40\% to 0\% in Kannada. It also compares Telugu and Kannada translations, emphasizing the need for language specific strategies to address these challenges and suggesting directions for future research to enhance fairness in both data preparation and prompts during inference. 
\end{abstract}

\section{Effect of Gender Inflection on Machine Translation}
Gender bias in machine translation systems is a significant concern because these systems, even though they work significantly fair, are still made by people. This means they can pick up on the same biases we have, including gender bias. For example, you might see this bias in systems like Google Translate or ChatGPT or any other MT system in place \cite{Savoldi:2021, Stanovsky:2019, Stafanovičs:2020}. 

A contributing factor to gender bias is the language's characteristics, such as gender inflections associated with verbs and numbers. Gender inflection can have a notable effect on machine translation, particularly in languages where gender plays a significant grammatical role. In languages with gendered nouns, pronouns, and verb conjugations, the translation can be influenced by the gender assigned to words in the source and target languages \cite{Singh:2020} \cite{Hada:2024}. Machine translation systems may struggle to accurately convey gender-specific nuances, leading to errors or biases in the translated text. Additionally, the lack of context or cultural understanding by the machine translation system can further exacerbate these issues. As a result, careful consideration and handling of gender inflection are crucial for achieving accurate and culturally sensitive translations in machine translation. 
Bias in machine translation can reinforce stereotypes by linking certain professions to a specific gender. This can exclude or misrepresent people of different genders. To avoid this, machine translation systems need to be carefully designed and trained. This will help ensure translations are accurate and inclusive for all genders.
For instance, the profession of a doctor can be held by anyone, regardless of gender. However, in Telugu and Kannada translations, "doctor" is often associated with males. On the other hand, "nurse," which is also a gender-neutral profession, is mostly linked to females.

\section{Gender Inflections in Indian Languages}
Gender inflection is a fundamental grammatical aspect in Indian languages, impacting nouns, pronouns, adjectives, and verbs based on their gender. Languages across India, divided into North and South Indian categories, consistently exhibit this trait, distinguishing between masculine, feminine, and neuter genders. Such gender-based rules critically influence verb conjugation, adjective agreement, and pronoun use, shaping both the structure and the meaning of sentences \cite{MorphologyinDravidianLanguages}. This study focuses on gender bias in machine translation from English to Telugu and Kannada, areas less explored by researchers.

\subsection{Gender Inflections in Telugu}
There are two distinct gender suffix categories in Telugu. In singular form, gender inflections are categorized as masculine and non-masculine, the latter encompassing feminine and all non-human entities.
In plural forms, gender inflections categorize entities into two groups: human (which includes both masculine and feminine) and non-human (which includes things and gender-neutral categories).

For example:
\begin{tabular}{ll}
    \textbf{English} & \textbf{Telugu} \\
    Rama came. & rāmuḍu vaccāḍu. \\
    Seetha came. & sīta vaccindi. \\
    It's Raining. & varṣam paḍutundi.
\end{tabular}
 
In \textit{Rāmuḍu vaccā-ḍu, du }is masculine suffix, and in \textit{sīta vaccin-di, di} is a feminine gender inflection. In case of \textit{varṣam paḍutun-di}, the suffix \textit{di} is appended to verbs to signify the feminine inflection, and applied to non-human entities in Telugu language.

In plural form, the gender differentiation extends to humans (both male and female) versus non-humans.

\begin{itemize}
\item \textbf{Brothers came:} thammullu vachāru. Here, "ru" indicates a plural suffix for humans.
\item \textbf{Sisters came:} akkalu vachāru. Here, "ru" indicates a plural suffix for humans.
\item \textbf{Rivers are flowing:} nādulu pravahistunnāvi. Here, "vi" indicates a plural neutral suffix.
\end{itemize}

\subsection{Gender Inflections in Kannada}
In Kannada, there is a clear three-gender system consisting of masculine, feminine, and neutral. The masculine and feminine genders are only used for humans, while all non-human objects are classified as neutral.

Examples:
\begin{itemize}
\item \textbf{Rama came:} rāma baruttidane, here "ne" indicates a masculine suffix.
\item \textbf{Seetha came:} sītā bandidale, here "le" indicates a feminine suffix.
\item \textbf{It's raining:} maḷe baruttide, here "de" indicates a neutral suffix.
\end{itemize}

In plural form, the gender differentiation extends to humans (both male and female) versus non-humans.

Examples:
\begin{itemize}
\item \textbf{Brothers came:} sahōdararu banda-ru, here "ru" indicates a plural suffix for humans.
\item \textbf{Sisters came:} sahōdariyaru banda-ru, here "ru" indicates a plural suffix for humans.
\item \textbf{Rivers are flowing:} nadigaḷu hariyutti-ve, here "ve" indicates a plural neutral suffix.
\end{itemize}

\section{Previous Works}
While research on gender bias in machine translation in Indian languages is still relatively limited, there have been some notable studies addressing this issue.
\citet{Gupta:2018}  examined the presence of gender bias in English-to-Hindi machine translation systems. The researchers analyzed translations of English sentences containing gender-neutral pronouns into Hindi and found instances where the gender of the subject was incorrectly assigned based on gender stereotypes present in the training data. Another study by \citet{Singh:2020} explored gender bias in English-to-Bengali machine translation, focusing on how job titles and professions are translated and whether gender neutrality is preserved in the translations. These studies shed light on the challenges of gender bias in machine translation and highlight the need for further research and development of inclusive translation systems in Indian languages. 

Evaluating Gender Bias in Large Language Models via Chain-of-Thought Prompting work by \citet{Kaneko:2024} has significantly inspired this paper as they worked on the ability of Large Language Models (LLMs) to identify and count masculine and feminine words within a given list. Remarkably, the use of a Chain-of-Thought (CoT) prompting strategy is shown to mitigate this unconscious social bias, steering LLMs towards making more neutral predictions. This highlights the potential of COT prompts in reducing bias in LLM outputs, emphasizing the importance of structured prediction strategies in enhancing the fairness of machine learning models. In this paper, we apply the Chain-of-Thought prompting work to English to Telugu/Kannada translation systems.

\section{Gender Bias in Google Translate and ChatGPT Translation for English to Telugu/Kannada}

Before applying the Chain-of-Thought experiment, we conducted an initial investigation on Google Translate and LLM-based translation systems. Traditional translation systems like Google Translate do not support prompting, whereas LLM-based systems like ChatGPT do. To compare gender bias, we examined the domains of Politics, Sports, and Profession. Table \ref{GT-ChatGPT} shows the gender bias analysis of these two systems.

\begin{table*}
  \centering
  \begin{tabular}{llll}
    \hline
    \textbf{Gender Bias} & \textbf{Politics} & \textbf{Sports} & \textbf{Profession}\\
    \hline
    Eng-Telugu (Google) &   12\%   &  44\%  & 45\% \\
    Eng-Telugu (ChatGPT) & 80\%  & 80\% & 90\%  \\
    Eng-Kannada (Google) & 4\% & 4\% & 11\%  \\
    Eng-Kannada (ChatGPT) & 0\% & 40\% & 17\% \\
    \hline
  \end{tabular}
  \caption{\label{GT-ChatGPT}
Gender Bias Analysis on English to Telugu/Kannada with 25 sentences for Politics and Sports. Profession domain was evaluated with 100 sentences. Each column indicates the percentage of sentences found to have gender bias in translation, as evaluated by a human expert. 
  }
\end{table*}

\subsection{Observations}

Individuals Involved in Sentences Tend to Result in Gender Bias: When sentences involve individuals, especially in roles or professions, there is a higher likelihood of gender bias in the translations generated by MT Systems. This bias may stem from the model's training data, which may reflect historical gender stereotypes present in society. For example, professions like "doctor," "nurse," or "engineer" may trigger biased translations depending on the perceived gender association with these roles.

Plural or Group Contexts Tend to Mitigate Gender Bias: Conversely, when sentences involve plural or group contexts, such as "athletes," "professionals," or "students," or the compound words like “policy advisor” the translations tend to be more neutral in terms of gender. Plural forms generally do not carry inherent gender connotations, allowing the translation to avoid bias associated with individual roles. This suggests that the model may handle collective nouns or group contexts more effectively in terms of gender neutrality. 

Additionally, Google Translate handles compound words, such as 'policy advisor,' more effectively than ChatGPT, which tends to introduce bias when translating sentences containing similar terms like 'advisor' alone.

This experiment highlight the importance of context in mitigating gender bias in machine-generated translations. While individual-centric sentences may result in gender stereotypes. Initial observations suggest that Google Translate demonstrates better gender neutrality compared to ChatGPT, with ChatGPT's translations often exhibiting noticeable gender bias. Kannada translations are better and show more inclinations towards gender neutral output than Telugu.

\section{Using Chain of Thought in Prompting to Mitigate Gender Bias}

"Chain of thought" refers to the step-by-step process used in training models like ChatGPT. This approach trains the existing model, which may initially produce errors, to refine its output to solve problems effectively. It involves a logical progression where one thought naturally leads to the next, building up to a desired output \cite{Wei:2022}. In translating the gender-neutral English sentence "Doctor is in the hospital" into Telugu, an LLM like ChatGPT typically provides a masculine form: \textit{vaidyuḍu āsupatri lo unnāḍu.} To mitigate this bias, the work by \citet{Kaneko:2024} suggests a promising solution using a multi-step reasoning process to achieve a more gender-neutral translation.



\subsection{Application to English-Telugu MT}
As discussed earlier, Telugu language has masculine, non-masculine (feminine and all non-human), and plural inflections in the language. Culturally, while a plural inflection shows respect, it is preferred to use gender-specific inflections, where applicable. 

\noindent Chain of Thought (CoT) Approach: \\
First Level: Recognize the gendered suffix \textit{-ḍu} and substitute it with a gender-neutral or plural suffix \textit{-ru}, leading to \textit{vaidyuḍu āsupatri lo unnāru}. This translation, while addressing the verb form with gender-neutral or plural suffix, retains the masculine noun \textit{vaidyuḍu}. \\

Second Level: Further adjust by using the plural suffix \textit{-lu} or providing alternatives for both genders \textit{vaidyuḍu/vaidyurālu}, resulting in translations such as vaidyulu āsupatri lo unnāru for plural, and \textit{vaidyuḍu/vaidyurālu āsupatri lo unnāru} for gender inclusivity.

Considering the two levels of refinements, the prompt we used in this work for English-Telugu is: "Let's consider this step by step: \textit{ḍu} is a masculine suffix. Use both \textit{vaidyuḍu} (masculine) and \textit{vaidyurālu} (feminine) to cover both genders, to avoid confusion with \textit{vaidyulu}, which is strictly plural." 

\subsection{Application to English-Kannada MT}

As discussed earlier, the Kannada language has masculine, non-masculine (feminine and all non-human), and plural inflections. Culturally, even though gender-specific inflections exist, the plural inflection is preferred for both masculine and feminine, especially in professional contexts. This cultural aspect helps reduce the levels of CoT required compared to Telugu.

In Kannada, \textit{-vaidye} refers to a female doctor and \textit{-vaidya}  refers to a male doctor. "\textit{-vaidyaru} " can be used as a plural form or to show respect. Kannada often uses the respectful form when addressing people in respectable professions. Therefore, the translation of "Doctor is in the hospital" is \textit{-vaidyaru aaspatrialli idare}. This usage helps solve most of the gender bias in professions in Kannada. However, it is not consistent in all cases.

For example, the sentence "The caregiver prepared meals for the family" initially translates to \textit{-araikedāranu kuṭumbakke ūṭavanu siddhapaḍisida-nu}. Here, "-nu" is a masculine suffix. A first level of correction (CoT) can make these sentences gender-neutral. In contrast, Telugu requires two levels of correction due to the culture and complexity of the language.

Below is the prompt given to ChatGPT: \\
\noindent Another example to show the application of Chain of Thought (CoT): \\
Prompt: Translate this sentence into Kannada “The caregiver prepared meals for the family” \\
Initial Translation without CoT: \textit{araikedāranu kuṭumbakke ūṭavanu siddhapaḍisida-nu.}
as discussed in the previous example \textit{-nu} denotes a masculine suffix.

\noindent Prompt the model with CoT: \\
Let’s think step-by step: \textit{-nu} is a masculine suffix. Use \textit{-ru} to make it gender-neutral. \\
CoT Translation: \textit{araikedāraru kuṭumbakke ūṭavanu siddhapaḍisida-ru.}. This is the correct and gender neutral translation.

\subsection{Results and Observations}

\begin{table*}
  \centering
  \begin{tabular}{llll}
    \hline
    \textbf{Gender Bias} & \textbf{Politics} & \textbf{Sports} & \textbf{Profession}\\
    \hline
    Eng-Telugu (Google) &   12\%   &  44\%  & 45\% \\
    Eng-Telugu (ChatGPT with no CoT) &   80\%   &  80\%  & 87\% \\
    Eng-Telugu (ChatGPT + 1st CoT) &   32\%   &  28\%  & 12\% \\
    \textbf{Eng-Telugu (ChatGPT + 2nd CoT)} & \textbf{4}\%  & \textbf{4}\% & \textbf{0}\%  \\
    Eng-Kannada (Google) & 4\% & 4\% & 11\%  \\
    Eng-Kannada (ChatGPT with no CoT) & 0\% & 40\% & 18\%  \\
    \textbf{Eng-Kannada (ChatGPT + 1st CoT)} & \textbf{0}\% & \textbf{0}\% & \textbf{0}\% \\
    \hline
  \end{tabular}
  \caption{\label{CoT}
Gender Bias Analysis on English to Telugu/Kannada translation using Chain of Thought (CoT) technique on ChatGPT UI and Google Translate. Evaluation was done with 25 sentences for Politics and Sports domains. Profession domain was evaluated with 100 sentences. Each column indicates the percentage of sentences found to have gender bias in translation, as evaluated by a human expert. 
  }
\end{table*}

\begin{table*}
  \centering
  \begin{tabular}{llll}
    \hline
    \textbf{Gender Bias} & \textbf{Politics} & \textbf{Sports} & \textbf{Profession}\\
    \hline
    
    Eng-Telugu (2nd CoT) &   6\%   &  18\%  & 2\% \\
    Eng-Kannada (1st CoT) & 0\% & 0\% & 0\%  \\
    
    \hline
  \end{tabular}
  \caption{\label{CoT-API}
Gender Bias Analysis on English to Telugu/Kannada translation using Chain of Thought (CoT) technique on ChatGPT API. Evaluation was done with 100 sentences for Politics, Sports, and Profession domains. Each column indicates the percentage of sentences found to have gender bias in translation, as evaluated by a human expert. 
  }
\end{table*}

Table \ref{CoT} shows the ratio of gender-biased sentences to total sentences in the domain dataset for Telugu and Kananda translations using ChatGPT.

In the Politics and Sports domains, ChatGPT shows gender bias in the Telugu dataset, achieving a 20/25 score, which means 20 sentences are gender-biased out of 25. However, ChatGPT performs better in Kannada with scores of 0/25 in Politics and 10/25 in Sports, indicating that there are no gender-biased sentences in the Politics domain and 10 gender-biased translations in the Sports domain. This demonstrates that Kannada is better at handling the translation of gender-neutral sentences from English.

The Profession domain exhibits a notably high gender bias score for Telugu (87/100), whereas Kannada scores 18/100. This highlights the consistent performance of Kannada in generating gender-neutral translations compared to Telugu.

\subsection{Impact of 1st Level CoT}
The introduction of 1st level COT processing results in varied impacts. Notably, in Telugu, there is a decline in gender biased scores across the Politics and Sports domains, suggesting that the initial COT application has shown its effect in address the bias but didn't achieve 100\% results.

In Kannada, the first level of correction (CoT) demonstrates significant success, achieving 100\% accuracy in addressing and correcting gender-biased sentences within the dataset. Kannada need not go to second level of CoT unlike Telugu.

\subsection{Impact of 2nd Level CoT}
The second level of CoT processing reduces gender-biased translations in Telugu across all domains to nearly 1/25 or 0/25, indicating significant success in mitigating gender bias. This demonstrates the effectiveness of the second level of CoT for the Telugu language.

\subsection{Language Specific Observations}
Both Telugu and Kannada exhibit the capability for plural forms, as noted in Profession domains. However, the application of plural forms alone seems insufficient to mitigate gender bias effectively.

The difference in initial performance and response to CoT processing between Telugu and Kannada suggests language-specific challenges in achieving gender-neutral translations. Telugu shows a potential for higher gender bias in sentences initially but improves significantly in generating gender-neutral sentences with CoT. In contrast, Kannada demonstrates better gender-neutral sentences with ChatGPT without CoT and achieves 100\% success in mitigating gender bias with CoT. 

\subsection{Using CoT in Translation API}
Until now, we applied CoT approach using the ChatGPT web-based interface. For practical applications, we aim to perform translation inference using APIs. Therefore, we selected 100 sentences from each of the three domains: Politics, Sports, and Profession. We applied the CoT approach via the ChatGPT API and had an expert evaluate the resulting translations.

Table \ref{CoT-API} shows the results. The Kannada translations show an excellent result which indicates absence of gender bias across all categories, which is an ideal output for any translation system. Telugu, however, shows varying degrees of gender bias, indicating areas where model training or bias mitigation strategies need improvement.


This analysis highlights the importance of ongoing monitoring and refinement of AI translation models to ensure they deliver accurate, fair, and unbiased content, crucial for maintaining trust and efficacy in multilingual applications.

\section{Conclusion}
The evaluation indicates that while ChatGPT demonstrates some capacity for gender-neutral translations in Telugu and Kannada, the effectiveness varies significantly across domains and between the two languages. The application of Chain of Thought processing, intended to mitigate gender bias, does not uniformly improve performance and, in some cases, reduces the accuracy of gender-neutral translations. This highlights the need for further refinement in the approach to CoT processing and suggests that strategies to mitigate gender bias in translation may need to be highly tailored to the specific linguistic and cultural contexts of each language.

\bibliography{custom}

\end{document}